\documentclass{article}
\usepackage{ijcai18}
\usepackage{balance}
\usepackage{times}
\usepackage{comment}
\usepackage{xcolor}
\usepackage{soul}
\usepackage[utf8]{inputenc}
\usepackage[small]{caption}
\usepackage{amsfonts}
\usepackage{amsmath}
\usepackage{epsfig}
\usepackage{amsopn}
\usepackage{float}
\usepackage{epstopdf}

\usepackage{url}
\usepackage{booktabs} 
\usepackage{algorithmic}
\usepackage{colortbl}
\usepackage[ruled]{algorithm2e} 

\usepackage{caption}
\usepackage{subcaption}
\usepackage{multirow}
\usepackage{tabularx}
\usepackage{bm}
\usepackage{graphicx}
\usepackage{rotating}
\usepackage{mathrsfs}

\title{Multi-modality Sensor Data Classification with Selective Attention}

\author{
Xiang Zhang$^\dag$,
Lina Yao$^\dag$,
Chaoran Huang$^\dag$,
Sen Wang$^\S$, 
Mingkui Tan$^\P$,
Guodong Long$^\ddag$,
Can Wang$^\S$
\\
$^\dag$ University of New South Wales \\
$^\S$ Griffith University\\
$^\P$ South China University of Technology\\
$^\ddag$ University of Technology Sydney \\
xiang.zhang3@student.unsw.edu.au,
\{lina.yao, chaoran.huang\}@unsw.edu.au,\\
 \{sen.wan, can.wang\}@griffith.edu.au, mingkuitan@scut.edu.cn,
guodong.long@uts.edu.au \\
}

\begin{document}

\maketitle
\begin{abstract}
Multimodal wearable sensor data classification plays an important role in ubiquitous computing and has a wide range of applications in scenarios from healthcare to entertainment. However, most existing work in this field employs domain-specific approaches and is thus ineffective in complex situations where multi-modality sensor data are collected. Moreover, the wearable sensor data are less informative than the conventional data such as texts or images. In this paper, to improve the adaptability of such classification methods across different application domains, we turn this classification task into a game and apply a deep reinforcement learning scheme to deal with complex situations dynamically. Additionally, we introduce a selective attention mechanism into the reinforcement learning scheme to focus on the crucial dimensions of the data. This mechanism helps to capture extra information from the signal and thus it is able to significantly improve the discriminative power of the classifier. We carry out several experiments on three wearable sensor datasets and demonstrate the competitive performance of the proposed approach compared to several state-of-the-art baselines.
\end{abstract}

\section{Introduction}
\label{sec:introduction}
Nowadays, diverse categories of sensors can be found in various wearable devices. Such devices are now being widely applied in multiple fields, such as Internet of Things \cite{kamburugamuve2015framework,zhang2017converting}. As a result, massive multimodal sensor data are being produced continuously. The question that how we can deal with these data efficiently and effectively has become a major concern.

Compared to conventional sensor data such as images and videos, these data are naturally formed as a 1-D signal, with each element representing one sensor channel accordingly. There are several challenges for such sensor data classification. First, most existing classification methods use domain-specific knowledge and thus may become ineffective or even fail in complex situations where multimodal sensor data are being collected \cite{bigdely2015prep}. For example, one approach that works well on IMU (Inertial Measurement Unit) signals may not be able to deal with EEG (Electroencephalography) brain signals. Therefore, an effective and universal sensor data classification method is highly desirable for complex situations. This framework is expected to have both efficiency and robustness over various sensor signals.

Second, the wearable sensor data carries far less information than texts and images. For example, a sample signal gathered by a 64-channel EEG equipment only contains 64 numerical elements. Hence, a more effective classifier is required to extract discriminative information from such limited raw data. However, maximizing the utilization of the given scarce data demands cautious preprocessing and a rich fund of domain knowledge.

Inspired by attention mechanism \cite{cavanagh1992attention}, we propose to concentrate on a focal zone of the signal to automatically learn the informative attention patterns for different sensor combinations. Here, the focal zone is a selection block of the signal with a certain length, sliding over the feature dimensions. Note that reinforcement learning has been shown to be capable of learning human-control level policy on a variety of tasks \cite{mnih2015human}. Then we exploit the reinforcement learning to discover the focal zone.
 Moreover, considering that the signals in different categories may have different inter-dimension dependency \cite{markham1981land}, we propose to use the LSTM (Long Short-Term Memory \cite{gers1999learning,zhang2017eeg}) to exploit the latent correlation between signal dimensions. We propose a weighted average spatial LSTM (WAS-LSTM) classifier by exploring the dependency in sensor data.

The main contributions of this paper are as follows:
\begin{figure*}[!t]
\centering
  \includegraphics[width=0.9\linewidth]{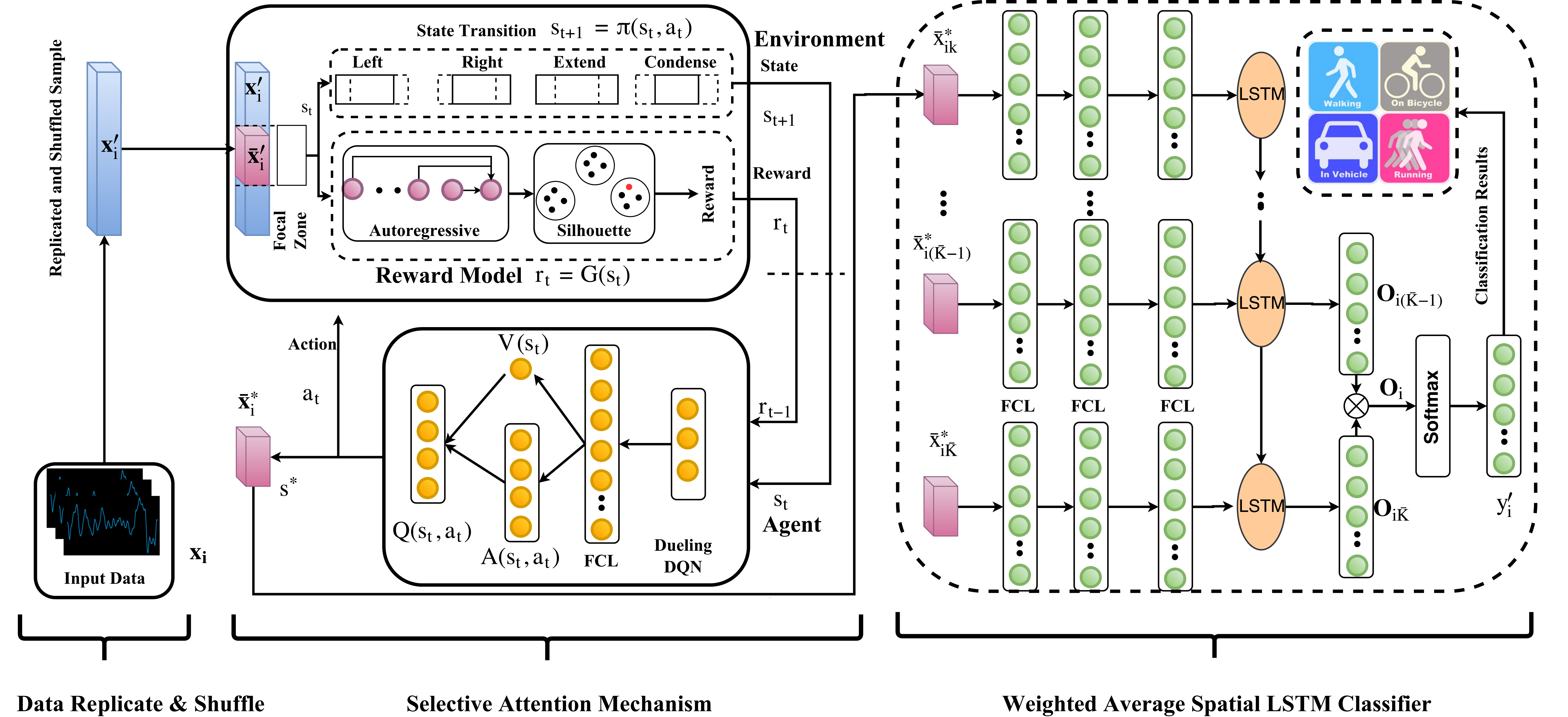}
  \caption{Flowchart of the proposed approach.
   The focal zone $\mathbf{\bar{x}}_i$ is a selected fragment from  $\mathbf{x}'_i$ to feed in the state transition and the reward model.
   In each step $t$, one action is selected by the state transition to update $s_t$ based on the agent's feedback. The reward model evaluates the quality of the focal zone to the reward $r_t$.
   The dueling DQN is employed to find the optimal focal zone $\mathbf{\bar{x}}^*_i$ which will be feed into the LSTM based classifier to explore the inter-dimension dependency and predict the sample's label $y'_i$. $FCL$ denotes Fully Connected Layer. The State Transition contains four actions: left shifting, right shifting, extend, and condense. The dashed line indicates the focal zone before the action while the solid line indicates the position of the focal zone after the action.}
  \label{fig:workflow}
\end{figure*}

\begin{itemize}
  \item We propose a selective attention mechanism for
sensor data classification using the spatial information only. The proposed method is insensitive to sensor types since it is capable of handling multimodal sensor data.
\item  We apply deep reinforcement learning to automatically select the most distinguishable features, called focal zone, for multimodal sensor data of different sensor types and combinations.  We design a novel objective function as the award in reinforcement learning task to optimize the focal zone. The new reward model saves more than 98\% training time of the deep reinforcement learning.
\item We propose Weighted Average Spatial LSTM classifier to capture the cross-dimensional dependency in multimodal sensor data. 
\end{itemize}

\section{Proposed Method}
\label{sec:methodology}

Suppose the input sensor data can be denoted by $\mathbf{X}=\{(\mathbf{x}_i, y_i), i=1,2,\cdots I\}$ where $(\mathbf{x}_i, y_i)$ denotes the 1-D sensor signal, called one \textit{sample} in this paper, and $I$ denotes the number of samples. In each sample, the feature $\mathbf{x}_i\in \mathbb{R}^K$ contains $K$ elements and the corresponding ground truth $y_i\in \mathbb{R}$ is an integer denotes the sample's category. $\mathbf{x}_i$ can be described as a vector with $K$ elements, $\mathbf{x}_i=\{x_{ik},k=1,2,\cdots, K\}$.

The proposed algorithm is shown in Figure~\ref{fig:workflow}. The main focus of the algorithm is to exploit the latent dependency between different signal dimensions. To this end, the proposed approach contains several components: 1) the replicate and shuffle processing; 2) the selective attention learning; 3) the sequential LSTM-based classification. In the following, we will first discuss the motivations of the proposed method and then introduce the aforementioned components in details.

\subsection{Motivation} 
\label{sub:motivation}
How to exploit the latent relationship between sensor signal dimensions is the main focus of the proposed approach. The signals belonging to different categories are supposed to have different inter-dimension dependent relationships which contain rich and discriminative information. This information is critical to improve the distinctive signal pattern discovery.

In practice, the sensor signal is often arranged as 1-D vector, the signal is less informative for the limited and fixed element arrangement. The elements order and the number of elements in each signal vector can affect the element dependency.
In many real-world scenarios, the multimodal sensor data are associated with the practical placement. For example, the EEG data are concatenated following the distribution of biomedical EEG channels.
Unfortunately,  the practical sensor sequence, with the fixed order and number, may not be suitable for inter-dimension dependency analysis. Meanwhile, the optimal dimension sequence \cite{tan2015lstm} varies with the sensor types and combinations. Therefore, we propose the following three techniques to amend these drawbacks.

     First, we replicate and shuffle the input sensor signal vector on dimension-wise in order to provide as much latent dependency as possible among feature dimensions (Section~\ref{sub:repeat_and_shuffle}).

     Second, we introduce a focal zone as a selective attention mechanism, where the optimal inter-dimension dependency for each sample only depends on a small subset of features. Here, the focal zone is optimized by deep reinforcement learning which has been proved to be stable and well-performed in policy learning (Section~\ref{sub:attention_pattern_learning}).

     Third, we propose the WAS-LSTM classifier by extracting the distinctive inter-dimension dependency (Section~\ref{sub:classification}).

\subsection{Data Replicate and Shuffle} 
\label{sub:repeat_and_shuffle}

To provide more potential inter-dimension spatial dependencies, we propose a method called Replicate and Shuffle (RS). RS is a two-step feature transformation method which maps $\mathbf{x}_i$ to a higher dimensional space $\mathbf{x}'_i$ with more complete element combinations:
$$\mathbf{x}_i\in \mathbb{R}^K \rightarrow \mathbf{x}'_i \in \mathbb{R}^{K'}, K'>K$$
In the first step (Replicate), replicate $\mathbf{x}_i$ for $h = K'\%K+1$ times where $\%$ denotes remainder operation. Then we get a new vector with length as $h*K$ which is not less than $K'$; in the second step (Shuffle), we randomly shuffle the replicated vector in the first step and intercept the first $K'$ element to generate $\mathbf{x}'_i$. Theoretically, compared to $\mathbf{x}_i$, $\mathbf{x}'_i$ contains more diverse and complete inter-dimension dependencies.

 \subsection{Selective Attention Mechanism} 
 \label{sub:attention_pattern_learning}
In the next process, we attempt to find the optimal dependency which includes the most distinctive information. But $K'$, the length of $\mathbf{x}'_i$, is too large and is computationally expensive. To balance the length and the information content, we introduce the attention mechanism \cite{cavanagh1992attention}.
We introduce the attention mechanism to emphasize the informative fragment in $\mathbf{x}'_i$ and denote the fragment by $\mathbf{\bar{x}}_i$, which is called \textit{focal zone}.
Suppose $\mathbf{\bar{x}}_i \in \mathbb{R}^{\bar{K}}$ and $\bar{K}$ denotes the length of the focal zone. For simplification, we continue denote the $k$-th element by $\mathbf{\bar{x}}_{ik}$ in the focal zone. To optimize the focal zone, we employ deep reinforcement learning as the optimization framework for its excellent performance in policy optimization \cite{mnih2015human}.

\textbf{Overview.} As shown in Figure~\ref{fig:workflow}, the focal zone optimization includes two key components: the environment (including state transition and reward model), and the agent. Three elements (the state $s$, the action $a$, and the reward $r$) are exchanged in the interaction between the environment and the agent. All of the three elements are customized based on the specific situation in this paper. Next, we introduce the design of the crucial components of our deep reinforcement learning structure:
\begin{itemize}
    \item The \textbf{state} $\mathcal{S}=\{s_t, t=0,1,\cdots,T\}\in \mathbb{R}^2$ describes the position of the focal zone, where $t$ denotes the time stamp. In the training, $s_0$ is initialized as $s_0=[(K'-\bar{K})/2, (K'+\bar{K})/2]$. Since the focal zone is a shifting fragment on 1-D $\mathbf{x}'_i$, we design two parameters to define the state: $s_t = \{start^t_{idx},end^t_{idx}\}$, where $start^t_{idx}$ and $end^t_{idx}$ separately denote the start index and the end index of the focal zone\footnote{For example, for a random $\mathbf{x}'_i = [3,5,8,9,2,1,6,0]$, the state $\{start^t_{idx}=2, end^t_{idx}=5\}$ is sufficient to determine the focal zone as $[8,9,2,1]$.}.
    \item The \textbf{action} $\mathcal{A}=\{a_t,t=0,1,\cdots,T\}\in \mathbb{R}^4$ describes which the agent could choose to act on the environment. In our case, we define 4 categories of actions for the focal zone (as described in the \textbf{State Transition} part in Figure~\ref{fig:workflow}): left shifting, right shifting, extend, and condense. Here at time stamp $t$, the state transition only choose one action to implement following the agent's policy $\pi$: $s_{t+1}=\pi(s_t,a_t)$.
    \item The \textbf{reward} $\mathcal{R}=\{r_t,t=0,1,\cdots,T\}\in \mathbb{R}$ is calculated by the reward model, which will be detailed later. The reward model $\Phi$: $r_{t}=\Phi(s_t)$ receives the current state and returns an evaluation as the reward.
    \item We employ the Dueling DQN (Deep Q Networks \cite{wang2015dueling}) as the optimization \textbf{policy} $\pi(s_t,a_t)$, which is enabled to learn the state-value function efficiently. Dueling DQN learns the Q value $V(s_t)$ and the advantage function $A(s_t,a_t)$ and combines them: $Q(s_t, a_t)\leftarrow V(s_t), A(s_t,a_t)$.
    The primary reason we employ a dueling DQN to optimize the focal zone is that it updates all the four Q values\footnote{Since we have four actions in $a_t$, the $Q(s_t, a_t)$ contains 4 Q values. The arrangement is similar with the one-hot label.} at every step while other policy only updates one Q value at each step.
\end{itemize}

\textbf{Reward Model.} Next, we detailedly introduce the design of the reward model for it is one crucial contribution of this paper. The purpose of reward model is to evaluate how the current state impact our final target which refers to the classification performance in our case. Intuitively, the state which can lead to the better classification performance should have a higher reward: $r_t=\mathcal{F}(s_t)$. As a result, in the standard reinforcement learning framework, the original reward model regards the classification accuracy as the reward. $\mathcal{F}$ refers to the WAS-LSTM.
Note, WAS-LSTM focuses on the spatial dependency between different dimensions at the same time-point while the normal LSTM focuses on the temporal dependency between a sequence of samples collected at different time-points.
However, the WAS-LSTM requires considerable training time, which will dramatically increase the optimization time of the whole algorithm. In this section, we propose an alternative method to calculate the reward: construct a new reward function $r_t=\mathcal{G}(s_t)$ which is positively related with $r_t=\mathcal{F}(s_t)$. Therefore, we can employ $\mathcal{G}$ to replace $\mathcal{F}$. Then, the task is changed to construct a suitable $\mathcal{G}$ which can evaluate the inter-dimension dependency in the current state $s_t$ and feedback the corresponding reward $r_t$. We propose an alternative $\mathcal{G}$ composed by three components: the autoregressive model \cite{akaike1969fitting} to exploit the inter-dimension dependency in $\mathbf{x}'_i$, the Silhouette Score \cite{laurentini1994visual} to evaluate the similarity of the autoregressive coefficients, and the reward function based on the silhouette score.

The autoregressive model \cite{akaike1969fitting} receives the focal zone $\mathbf{\bar{x}}_i$ and specifies that how the last variable depends on its own previous values.
$$\bar{x}_{i\bar{K}} = \sum_{j=1}^{p}\varphi_j \bar{x}_{i(\bar{K}-j)}+ C +\bar{\varepsilon}$$
where $p$ is the order of the autoregressive model, $C$ indicates a constant, and $\bar{\varepsilon}$ indicates the withe noise. From this equation, we can infer that
the autoregressive coefficient $\boldsymbol{\varphi}=\{\varphi_j,j=1,2,\cdots,p\} \in \mathbb{R}^p$ incorporates the dependent relationship in the focal zone. Then, to evaluate how rich information is taken in the $\boldsymbol{\varphi}$, we employ silhouette score \cite{lovmar2005silhouette}
to interpret the consistence of $\boldsymbol{\varphi}$. The higher score $ss_t$ indicates the focal zone in state $s_t$ contains more inter-dimension dependency, which means $\mathbf{\bar{x}}_i$ is easier to be classified by the classifier in the next stage. At last, based on the $ss_t\in[-1,1]$, we design a \textbf{reward function}:
$$r_t = \frac{e^{ss_t+1}}{e^2-1}-\beta \frac{\bar{K}}{K'}$$
The function contains two parts, the first part is a normalized exponential function with the exponent $ss_t+1 \in[0,1]$, this part encourages the reinforcement learning algorithm to search the better $s_t$ which leads to a higher $ss_t$. The motivation of the exponential function is that: the reward growth rate is increasing with the silhouette score's increase\footnote{For example, for the same silhouette score increment 0.1, $ss_t: 0.9\rightarrow 1.0$ can earn higher reward increment than $ss_t: 0.1\rightarrow 0.2$.}. The second part is a penalty factor for the focal zone length to keep the bar shorter and the $\beta$ is the penalty coefficient.

In summary, the aim of focal zone optimization is to learn
the optimal focal zone $\mathbf{\bar{x}}^*_i$ which can lead to the maximum reward. The optimization totally iterates $N=n_e*n_s$ times where $n_e$ and $n_s$ separately denote the number of episodes and steps \cite{wang2015dueling}. $\varepsilon$-greedy method \cite{tokic2010adaptive} is employed in the state transition.

\subsection{Weighted Average Spatial LSTM Classifier} 
\label{sub:classification}
In this section, we propose Weighted Average Spatial LSTM classification for two purposes. The first attempt is to capture the cross-relationship among feature dimensions in the optimized focal zone $\mathbf{\bar{x}}^*_i$. The LSTM-based classifier is widely used for its excellent sequential information extraction ability which is approved in several research areas such as natural language processing \cite{gers2001lstm,sundermeyer2012lstm}. Compared to other commonly employed spatial feature extraction methods, such as Convolutional Neural Networks, LSTM is less depends on the hyper-parameters setting. However, the traditional LSTM focuses on the temporal dependency among a sequence of samples. Technically, the input data of traditional LSTM is 3-D tensor shaped as $[n_b, n_t, \bar{K}]$ where $n_b$ and $n_s$ denote the batch size and the number of temporal sample, separately. The WAS-LSTM aims to capture the dependency among various dimensions at one temporal point, therefore, we set $n_t=1$ and transpose the input data as: $[n_b, n_t, \bar{K}]\rightarrow[n_b, \bar{K}, n_t]$.

The second advantage of WAS-LSTM is that it could stabilize the performance of LSTM via moving average method \cite{lipton2015learning}.
 Specifically, we calculate the LSTM outputs $\mathbf{O}_i$ by averaging the past two outputs instead of only the final one (Figure~\ref{fig:workflow}):
$$\mathbf{O}_i = (\mathbf{O}_{i(\bar{K}-1)}+\mathbf{O}_{i\bar{K}})/2$$
The predicted label is calculated by $y'_i = \mathcal{L}(\mathbf{\bar{x}}^*_i)$ where $\mathcal{L}$ denotes the LSTM algorithm. $\ell_2$-norm (with parameter $\lambda$) is adopted as regularization to prevent overfitting. The sigmoid activation function is used on hidden layers. The loss function is cross-entropy and is optimized by the AdamOptimizer algorithm \cite{kingma2014adam}.


\section{Experiments} 
\label{sec:experiments}
In this section, we evaluate the proposed approach over 3 sensor signal datasets (separately collected by EEG headset, environmental sensor, and wearable sensor) including 2 widely used public datasets and 2 limited but more practical local datasets. Firstly we describe the details of each dataset. Secondly, we demonstrate the effectiveness and robustness by comparing the performance of our approach to baselines and state-of-the-art. Lastly, we provide the efficiency of the alternative reward model designed in Section~\ref{sub:attention_pattern_learning}.

\begin{table}[]
\centering
\caption{Datasets description. PID denotes Person Identification, AR denotes Activity Recognition, and S-rate denotes Sampling rate. \#-S, \#-C, \#-D separately denote the number of subjects, classes, and dimensions.}
\label{tab:datasets}
\resizebox{\linewidth}{!}{
\begin{tabular}{llllllll}
\hline
\textbf{Datasets} & \textbf{Type} & \textbf{Task} & \textbf{\#-S} & \textbf{\#-C} & \textbf{Samples} & \textbf{\#-D} & \textbf{S-rate (Hz)} \\ \hline
\textbf{EID} & EEG & PID & 8 & 8 & 168,000 & 14 & 128 \\
\textbf{RSSI} & RFID & AR & 6 & 21 & 3,100 & 12 & 2 \\
\textbf{PAMAP2} & IMU & AR & 9 & 8 & 120,000 & 14 & 100 \\
\hline
\end{tabular}
}
\end{table}
\subsection{Datasets} 
\label{sub:datasets}
More details refer to Table~\ref{tab:datasets}.
\begin{itemize}
    \item \textbf{EID.} The EID (EEG ID identification) is collected in a constrained setting where 8 subjects (5 males and 3 females) aged $26\pm 2$. EEG signal monitors the electrical activity of the brain. This dataset gathers the raw EEG signals by Emotiv EPOC+ headset with 14 channels at the sampling rate of 128 Hz.
    \item \textbf{RSSI.} The RSSI (Radio Signal Strength Indicator) \cite{yao2018compressive} collects the signals from passive RFID tags. 21 activities, including 18 ADLs (Activity of Daily Living) and 3 abnormal falls, are performed by 6 subject aged $25\pm 5$. RSSI measures the power present in a received radio signal, which is a convenient environmental measurement in ubiquitous computing.
    \item \textbf{PAMAP2.} The PAMAP2 \cite{fida2015} is collected by 9 participants (8 males and 1 females) aged $27\pm 3$. 8 ADLs are selected as a subset of our paper. The activity is measured by 1 IMU attached to the participants' wrist. The IMU collects sensor signal with 14 dimensions including two 3-axis accelerometers, one 3-axis gyroscopes, one 3-axis magnetometers and one thermometer.
\end{itemize}

\begin{table}[t]
\centering
\caption{Comparison of EID}
\label{tab:comparison_eid}
\resizebox{\linewidth}{!}{
\begin{tabular}{cp{3cm}|llll}
\hline
\multicolumn{2}{c|}{\multirow{2}{*}{\textbf{Methods}}} & \multicolumn{4}{c}{\textbf{EID Dataset}} \\ \cline{3-6}
\multicolumn{2}{l|}{} & \textbf{Acc} & \textbf{Pre} & \textbf{Rec} & \textbf{F1} \\\hline
 & \textbf{SVM} & 0.1438 & 0.1653 & 0.1545 & 0.1445 \\
 & \textbf{RF} & 0.9365 & 0.9261 & 0.9142 & 0.9457 \\
 & \textbf{KNN} & 0.9413 & 0.9471 & 0.9298 & 0.9511 \\
 & \textbf{AB} & 0.2518 & 0.2684 & 0.2491 & 0.2911 \\
\multirow{-5}{*}{\textbf{\begin{tabular}[c]{@{}c@{}}Non-\\ DL\\\end{tabular}}} & \textbf{LDA} & 0.1485 & 0.1524 & 0.1358 & 0.1479 \\ \hline
 & \textbf{LSTM} & 0.4315 & 0.5132 & 0.4278 & 0.4532 \\
 & \textbf{GRU} & 0.4314 & 0.455 & 0.4288 & 0.4218 \\
\multirow{-3}{*}{\textbf{\begin{tabular}[c]{@{}c@{}}DL\\ \end{tabular}}} & \textbf{1-D CNN} & 0.8031 & 0.8127 & 0.805 & 0.8278 \\ \hline
\multicolumn{1}{l}{\textbf{}} & \textbf{WAS-LSTM} & 0.9518 & 0.9657 & \textbf{0.9631} & \textbf{0.9658} \\
\multicolumn{1}{l}{\textbf{}} & \textbf{Ours} & \textbf{0.9621} & \textbf{0.9618} & 0.9615 & 0.9615\\ \hline
\end{tabular}
}
\end{table}

\begin{table}[t]
\centering
\caption{Comparison of RSSI}
\label{tab:comparison_rssi}
\resizebox{\linewidth}{!}{
\begin{tabular}{cp{3cm}|llll}
\hline
\multicolumn{2}{c|}{\multirow{2}{*}{\textbf{Methods}}} & \multicolumn{4}{c}{\textbf{RSSI Dataset}} \\ \cline{3-6}
\multicolumn{2}{l|}{} & \textbf{Acc} & \textbf{Pre} & \textbf{Rec} & \textbf{F1} \\\hline
 & \textbf{SVM} & 0.8918 & 0.8924 & 0.8908& 0.8805 \\
 & \textbf{RF} & 0.9614 & 0.9713& 0.9652 & 0.9624 \\
 & \textbf{KNN} & 0.9612 & 0.9628 & 0.9618 & 0.9634 \\
 & \textbf{AB} & 0.4704 & 0.4125 & 0.4772 & 0.3708 \\
\multirow{-5}{*}{\textbf{\begin{tabular}[c]{@{}c@{}}Non-\\ DL\\ \end{tabular}}} & \textbf{LDA} & 0.8842 & 0.8908 & 0.8845 & 0.8802 \\ \hline
 & \textbf{LSTM} & 0.7421 & 0.6505 & 0.6132 & 0.6858 \\
 & \textbf{GRU} & 0.7049 & 0.7728 & 0.6584 & 0.6915 \\
\multirow{-3}{*}{\textbf{\begin{tabular}[c]{@{}c@{}}DL\\ \end{tabular}}} & \textbf{1-D CNN} & 0.9714 & 0.9676 & 0.9635 & 0.9645 \\ \hline
\multicolumn{1}{l}{\textbf{}} & \textbf{WAS-LSTM} & 0.9553 & 0.9533 & 0.9545 & 0.9592 \\
\multicolumn{1}{l}{\textbf{}} & \textbf{Ours} & \textbf{0.9838} & \textbf{0.9782} & \textbf{0.9669} & \textbf{0.9698}\\ \hline
\end{tabular}
}
\end{table}

\begin{table}[t]
\centering
\caption{Comparison of PAMAP2}
\label{tab:comparison_pamap2}
\resizebox{\linewidth}{!}{
\begin{tabular}{cp{3cm}|llll}
\hline
\multicolumn{2}{c|}{\multirow{2}{*}{\textbf{Methods}}} & \multicolumn{4}{c}{\textbf{PAMAP2 Dataset}} \\ \cline{3-6}
\multicolumn{2}{l|}{} & \textbf{Acc} & \textbf{Pre} & \textbf{Rec} & \textbf{F1} \\\hline
\multicolumn{1}{c}{} & \textbf{SVM} & 0.7492&  0.7451 & 0.7522 & 0.7486 \\
\multicolumn{1}{c}{} & \textbf{RF} & 0.9817 &0.9893 & 0.9711 & \textbf{0.9801} \\
\multicolumn{1}{c}{} & \textbf{KNN} & 0.9565 & 0.9651 & 0.9625&  0.9638 \\
\multicolumn{1}{c}{} & \textbf{AB} & 0.5776& 0.4298 & 0.5814 & 0.4942 \\
\multicolumn{1}{c}{\multirow{-5}{*}{\textbf{\begin{tabular}[c]{@{}c@{}}Non-\\ DL\end{tabular}}}} & \textbf{LDA} & 0.7127  &0.7175  &0.7298 & 0.7236 \\\hline
\multicolumn{1}{c}{} & \textbf{LSTM} & 0.7925 &0.7487 & 0.7478  &0.7482 \\
\multicolumn{1}{c}{} & \textbf{GRU} & 0.8625 & 0.8515 & 0.8349 & 0.8431 \\
\multicolumn{1}{c}{\multirow{-3}{*}{\textbf{\begin{tabular}[c]{@{}c@{}}DL\\\end{tabular}}}} & \textbf{1-D CNN} & 0.9819& 0.9715 & 0.9721&  0.9718 \\\hline
 & \textbf{\cite{fida2015}} & 0.96 & - & - & - \\
 & \textbf{\cite{chowd2017}} & 0.8488 & - & - & 0.841 \\
 & \textbf{\cite{erfani2017}} & 0.967 & - & - & - \\
\multirow{-4}{*}{\textbf{\begin{tabular}[c]{@{}l@{}}State-\\ of-the \\ -Arts\end{tabular}}} & \textbf{\cite{zheng2014time}} & 0.9336 & - & - & - \\\hline
\textbf{} & \textbf{WAS-LSTM} & 0.9821 & \textbf{0.9981} & 0.9459  &0.9713 \\
\textbf{} & \textbf{Ours} & \textbf{0.9882} & 0.9804 & \textbf{0.9756} & 0.9780 \\\hline
\end{tabular}
}
\end{table}
\vspace{-2mm}

\subsection{Results} 
\label{sub:comparision}
In this section, we compare the proposed approach with baselines and the state-of-the-art methods.
Our method focuses on the focal zone which is optimized by deep reinforcement learning and then explores the dependency between sensor signal elements by a deep learning classifier. All the three datasets are randomly split into the training set (90\%) and the testing set ($10\%$). Each sample is one sensor vector recording collected at one time point.
Through the previous experimental tuning and the Orthogonal Array based hyper-parameters tuning method \cite{zhang2017intent}, the hyper-parameters are set as following.
In the selective attention learning: the order of autoregressive is 3; $\bar{K}=128$, the Dueling DQN has 4 lays and the node number in each layer are: 2 (input layer), 32 (FCL), 4 ($A(s_t,a_t)$) + 1 ($V(s_t)$), 4 (output). The decay parameter $\gamma =0.8$, $n_e=n_s=50$, $N=2,500$, $\epsilon=0.2$, learning rate$ =0.01$, memory size $ =2000$, length penalty coefficient $\beta=0.1$, and the minimum length of focal zone is set as 10. In the deep learning classifier: the node number in the input layer equals to the number of feature dimensions, three hidden layers with 164 nodes, two layers of LSTM cells and one output layer. The learning rate $ =0.001$, $\ell_2$-norm coefficient $\lambda=0.001$, forget bias $=0.3$, batch size $ =9$, and iterate for 1000 iterations.

Tables~\ref{tab:comparison_eid} $\sim$~\ref{tab:comparison_pamap2} show the classification metrics comparison between our approach and baselines including Non-DL and DL baselines. Since the EID and RSSI are local datasets, we only compare with state-of-the-art over the public dataset PAMAP2. Table~\ref{tab:comparison_pamap2} shows that our approach achieves the highest accuracy on both datasets. DL represents deep learning. The notation and hyper-parameters of the baselines are listed here. RF denotes Random Forest, AdaB denotes Adaptive Boosting, LDA denotes Linear Discriminant Analysis. In addition, the key parameters of the baselines are listed here: Linear SVM ($C=1$), RF ($n=200$), KNN ($k=3$). In LSTM, $n_{steps}=5$, another set is the same as the WAS-LSTM classifier, along with the GRU (Gated Recurrent Unit \cite{chung2014empirical}). The CNN works on sensor data and contains 2 stacked convolutional layers (both with stride $[1,1]$, patch $[2,2]$, zero-padding, and the depth are 4 and 8, separately.) and followed by one pooling layer (stride $[1,2]$, zero-padding) and one fully connected layer (164 nodes). Relu activation function is employed in the CNN.
The results from Tables~\ref{tab:comparison_eid} $\sim$~\ref{tab:comparison_pamap2} show that:
\begin{itemize}
    \item Our approach outperforms all the baselines and the state-of-the-arts over all the local and public datasets ranging from EEG, RFID to wearable IMU sensors;
    \item The sensor spatial based WAS-LSTM classifier achieves a high-level performance, which indicates the method that extracting inter-dimension dependency for classification is effective;
    \item Our method (WAS-LSTM with focal zone) performs better than WAS-LSTM, which illustrates that the learned informative attention is effective.
\end{itemize}

\begin{figure*}[t]
\hspace{5mm}
    \centering
    \begin{subfigure}[t]{0.28\textwidth}
        \centering
        \includegraphics[width=\textwidth]{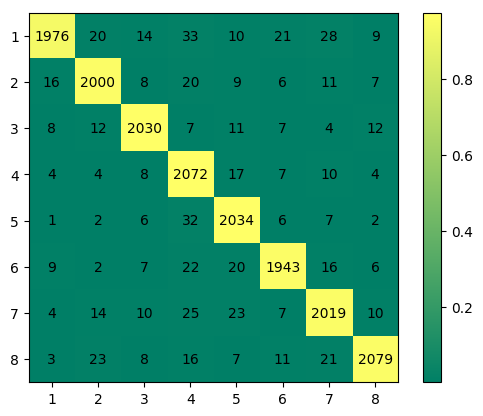}
        \caption{EID CM}
        \label{fig:eid_cm}
    \end{subfigure}%
    \begin{subfigure}[t]{0.28\textwidth}
        \centering
        \includegraphics[width=\textwidth]{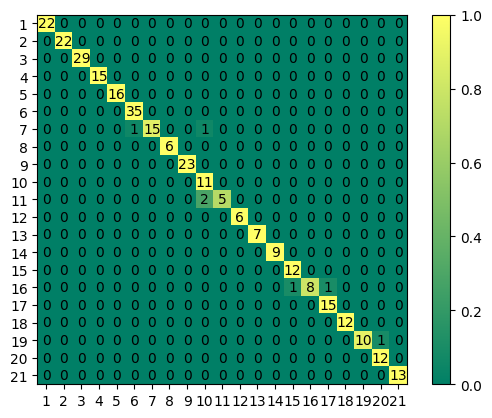}
        \caption{RSSI CM}
        \label{fig:rssi_cm}
    \end{subfigure}%
        \begin{subfigure}[t]{0.28\textwidth}
        \centering
        \includegraphics[width=\textwidth]{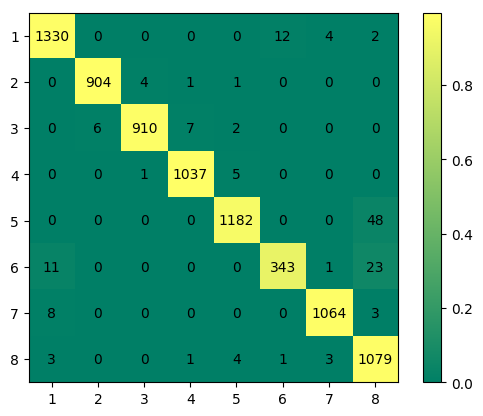}
        \caption{PAMAP CM}
        \label{fig:pamap_cm}    
    \end{subfigure}%

    \centering
    \begin{subfigure}[t]{0.28\textwidth}
        \centering
        \includegraphics[width=\textwidth]{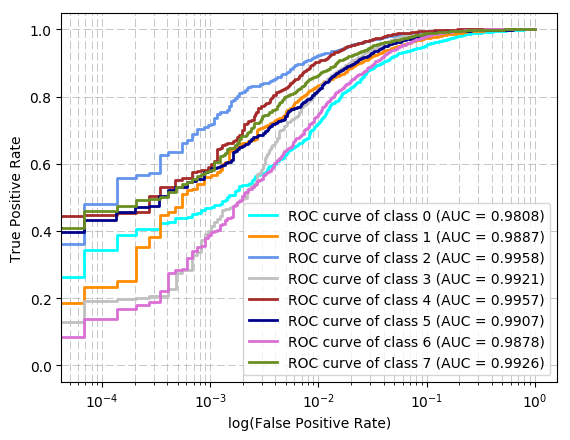}
        \caption{EID ROC}
        \label{fig:eid_roc}
    \end{subfigure}%
    \begin{subfigure}[t]{0.28\textwidth}
        \centering
        \includegraphics[width=\textwidth]{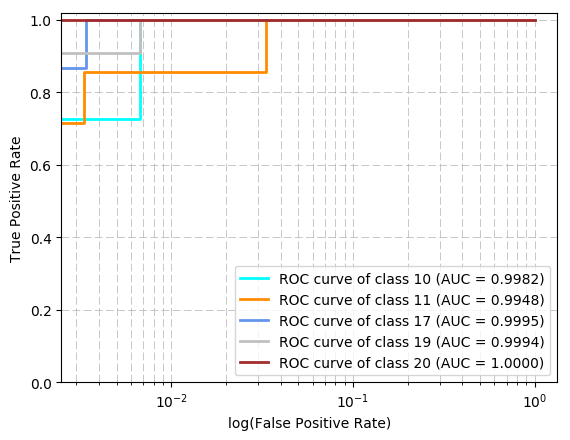}
        \caption{RSSI ROC}
        \label{fig:rssi_roc}
    \end{subfigure}%
        \begin{subfigure}[t]{0.28\textwidth}
        \centering
        \includegraphics[width=\textwidth]{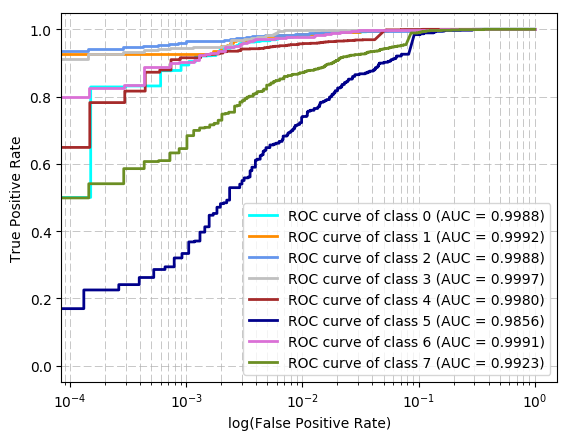}
        \caption{PAMAP ROC}
        \label{fig:pamap_roc}    
    \end{subfigure}%
    \caption{Confusion matrix and ROC curves of three datasets. CM denotes confusion matrix. The RSSI dataset overall contains 21 classes and we only select several representative classes.}
    \label{fig:cm_roc}
\end{figure*}

To have a closer observation, the CM (confusion matrix) and the ROC curves (including the AUC score) of the datasets are reported in Figure~\ref{fig:cm_roc}. The CMs illustrate that the robustness of the proposed approach keeps high accuracy even over few samples and numerous categories.

\subsection{Reward Model Efficiency Demonstration} 
\label{sub:reward_model_efficiency_demonstration}
In this paper, we propose a new reward model to replace the original reward model: $\mathcal{G} \rightarrow \mathcal{F}$. The original $\mathcal{F}$, in our case, refers to the WAS-LSTM classifier (Section~\ref{sub:classification}), intuitively. $\mathcal{F}$ requires a large amount of training time to find the optimal focal zone $\mathbf{\bar{x}^*}$. Take the EID dataset as an example, $\mathcal{F}$ needs around $4000$ sec on the Titan X (Pascal) GPU
for each step while the whole focal zone optimization contains $N$ ($N>2000$) iterations.
Therefore, to save training time, we attempt to employ $\mathcal{G}$
to approximate $\mathcal{G}$ to update the reward. Thus, two prerequisites are demanded: 1) $\mathcal{G}$ should have high correlation with $\mathcal{F}$ to guarantee $\mathop{\arg\max}\limits_{\mathbf{\bar{x}^*}}\mathcal{G} \approx  \mathop{\arg\max}\limits_{\mathbf{\bar{x}^*}}\mathcal{F}$; 2) the training time of $\mathcal{G}$ should be shorter than $\mathcal{F}$. In this section, we demonstrate the two prerequisites by experimental analyzes.

\begin{figure}[t]
\centering
\begin{minipage}[b]{0.44\linewidth}
\centering
\includegraphics[width=\textwidth]{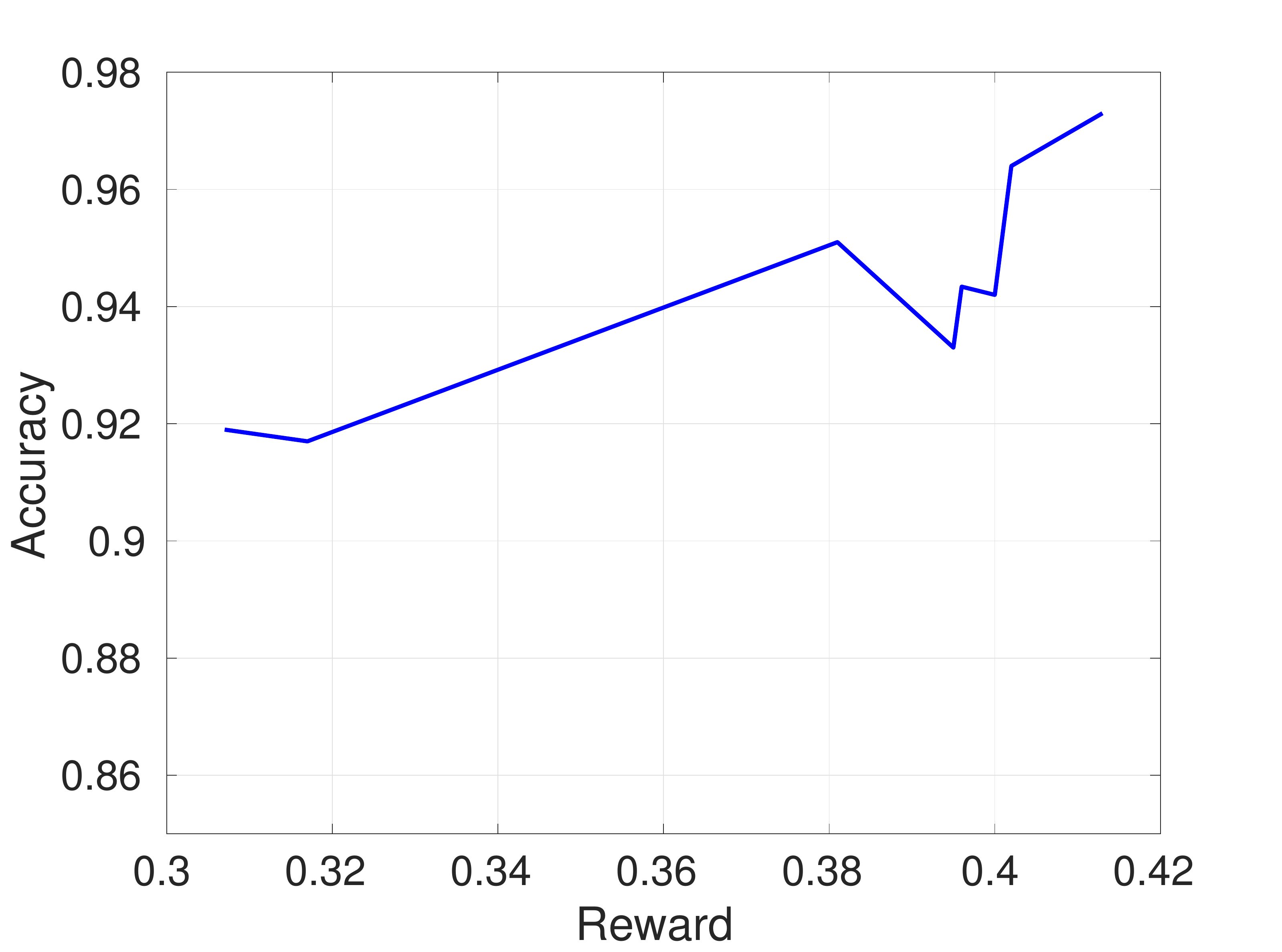}
\caption{The relationship between the classifier accuracy and the reward. The correlationship coefficient is 0.8258 while the p-value is 0.0115.}
\label{fig:reward_acc}
\end{minipage}
\hspace{1mm}
\begin{minipage}[b]{0.44\linewidth}
\centering
\includegraphics[width=\textwidth]{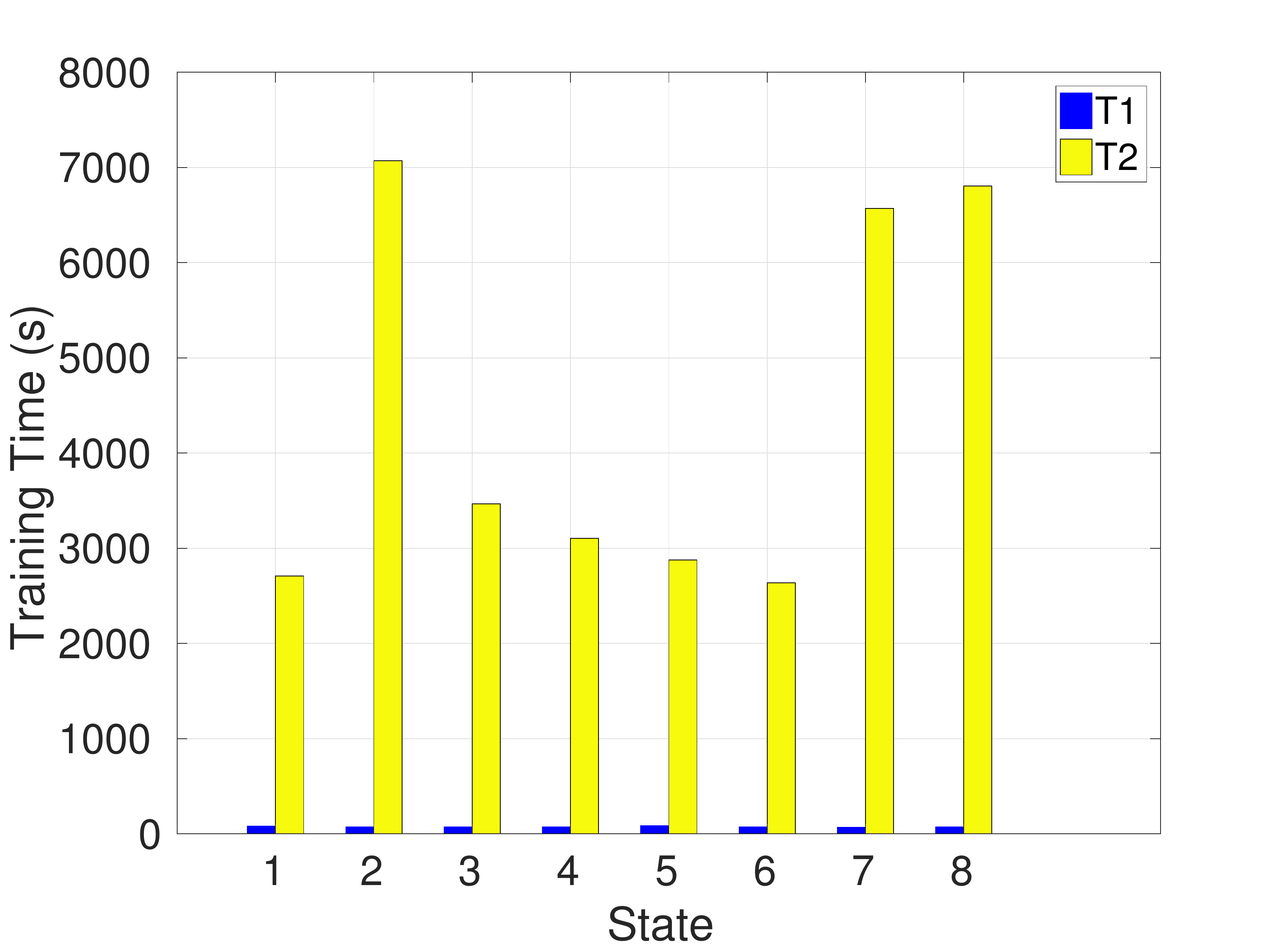}
\caption{Reward model training time in various states. T1 and T2 separately denote the training time in reward model $\mathcal{G}$ and $\mathcal{F}$. }
\label{fig:time_compare}
\end{minipage}
\end{figure}

First, on the focal zone optimization procedure on EID dataset, we conduct an experiment to measure a batch of data pairs of the reward (represents the reward of $\mathcal{G}$) and the WAS-LSTM classifier accuracy (represents the reward of $\mathcal{F}$). The relationship between the reward and the accuracy is shown in Figure~\ref{fig:reward_acc}. The figure illustrates that the accuracy has an approximately linear relationship with the reward. The correlations coefficient is 0.8258 (with p-value as 0.0115), which demonstrates the accuracy and reward are highly positive related. As a result, we can estimate $\mathop{\arg\max}\limits_{\mathbf{\bar{x}^*}}\mathcal{F}$ by $\mathop{\arg\max}\limits_{\mathbf{\bar{x}^*}}\mathcal{G}$. Moreover, another experiment is carried on to measure the single step training time of two reward models $\mathcal{G}$ and $\mathcal{F}$. The training times are marked as T1 and T2, respectively. Figure~\ref{fig:time_compare} qualitatively shows that T2 is much higher than T1 (8 states represent 8 different focal zones). Quantitatively, the sum of T1 over 8 states is $35237.41$ sec while the sum of T2 is $601.58$ sec. This results demonstrate that the proposed approach, designing a $\mathcal{G}$ to approximate and estimate the $\mathcal{F}$, saves $\mathbf{98.3\%}=1-601.58/35237.41$ training time in focal zone optimization.

\subsection{Discussions} 
\label{sub:discussion}
In this section, we discuss several characteristics of the proposed approach.

First, we propose a robust, universal, and adaptive classification framework which can efficiently deal with multi-modality sensor data. Specifically, our approach works better on high-dimensional feature space in that the information of inter-dimension dependency is richer.

In addition, we propose a novel idea that adopts an alternative reward model to estimate and replace the original reward model. In this way, the disadvantages of the original model, such as expensive computation, can be eliminated. The key is to keep the reward produced by the new model highly related to the original reward. The higher correlation coefficient, the better. This sheds light on the possible combination of deep learning classifier and reinforcement learning.

Nevertheless, one weakness is that the reinforcement learning policy only works well in the specific environment in which the model is trained. The dimension indices should be consistent in training and testing stages. Various policies should be trained according to different sensor combinations.

Furthermore, the proposed WAS-LSTM directly focuses on the dependency among the sensor dimensions and can produce a predicted label for each point.
This provides the foundation for the quick-reaction online detection and other applications which require instantaneous detection. However, this reuires a enough number of signal dimensions to carry sufficient information for the aim of accurately recognition.

\section{Conclusion} 
\label{sec:conclusion}
In this paper, we present a robust and efficient multi-modality sensor data classification framework
which integrates selective attention mechanism, deep reinforcement learning, and WAS-LSTM classification. In order to boost the chance of inter-dimension dependency in sensor features, we replicate and shuffle the sensor data. Additionally, the optimal spatial dependency is required for high-quality classification, for which we introduce the focal zone with attention mechanism. Furthermore, we extended the LSTM to exploit the cross-relationship among spatial dimensions, which is called WAS-LSTM, for classification. The proposed approach is evaluated on three different sensor datasets, namely, EEG, RFID and wearable IMU sensors. The experimental results show that our approach outperforms the state-of-the-art baselines. Moreover, the designed reward model saves $\textbf{98.3\%}$ of the training time in reinforcement learning.

\bibliographystyle{named}
\bibliography{ijcai18}

\end{document}